\documentclass[11pt,draftcls,onecolumn] {IEEEtran}
\ifCLASSINFOpdf
\else
\fi
\usepackage{amssymb,amsmath}
\usepackage[normalem]{ulem}
\usepackage[color,final]{showkeys}
\usepackage{multirow}
\usepackage{xhfill}
\usepackage{graphics}
\usepackage{graphicx}
\usepackage{subfigure}
\usepackage{epsfig}
\usepackage{url}

\begin{document}
%
\title{Randomized Structural Sparsity  based Support Identification with Applications to  Locating  Activated or Discriminative Brain Areas: A  Multi-center Reproducibility Study}
%
%
%

\author{Yilun~Wang, 
        Sheng~Zhang, 
        Junjie~Zheng,
        Heng~Chen,
        and~Huafu~Chen,  \thanks{This work was supported by the 973 project (Nos. 2015CB856000, 2012CB517901), 863 project (No. 2015AA020505), the Natural Science Foundation of China (Nos. 61125304, 11201054, 91330201), and the Specialized Research Fund for the Doctoral Program of Higher Education of China (No. 20120185110028), and the Fundamental Research Funds for the Central Universities (ZYGX2013Z004, ZYGX2013Z005). Corresponding authors: Yilun Wang and Huafu Chen.}
\thanks{Yilun~Wang and Sheng Zhang are with the School of Mathematical Sciences  and Center for Information in Biomedicine, University of Electronic Science and Technology of China, Chengdu,
Sichuan, 611731 P. R. China. (email: yilun.wang@rice.edu)
}
\thanks{Junjie~Zheng, Heng~Chen and  Huafu~Chen  are with Key laboratory for Neuroinformation of Ministry of Education,
School of Life Science and Technology, and Center for Information in Biomedicine,
University of Electronic Science and Technology of China,
Chengdu, Sichuan, 610054 P. R. China. (email: chenhf@uestc.edu.cn)}
}

%
%

\markboth{Journal of \LaTeX\ Class Files,~Vol.~11, No.~4, December~2012}%
{Shell \MakeLowercase{\textit{et al.}}: Bare Demo of IEEEtran.cls for Journals}
%



\maketitle

\begin{abstract}
In this paper, we focus on how to locate the relevant or discriminative brain regions related with external stimulus or certain mental decease, which is also called support identification, based on the neuroimaging data.
 The main difficulty lies in the extremely high dimensional voxel space and relatively few training samples, easily resulting in an unstable  brain region  discovery (or called feature selection in context of pattern recognition). When the training samples are from different centers and have between-center variations, it will be even harder to obtain a reliable and consistent result. 
Corresponding, we revisit our recently proposed algorithm based on  stability selection and
%
%
%
structural sparsity.  
It is applied to the multi-center MRI data analysis  for the first time. A consistent and stable result is achieved across different centers despite the between-center data variation while many other state-of-the-art
methods such as two sample t-test fail. Moreover, we have empirically showed that the performance of this algorithm is  robust and insensitive to several of its key parameters.
In addition, the support identification results on both functional MRI and structural MRI are interpretable and can be  the potential biomarkers.

\end{abstract}

%
\IEEEpeerreviewmaketitle

\section{Introduction} \label{Sec:intro}
\subsection{Problem Statement}
For computer multimedia analysis, a challenging problem is the semantic gap between the high-level perception and cognition  and the low-level features of the digital contents. Considering that the human brain can eliminate this gap very naturally, deep understanding of the brain responses to multimedia will be playing an important role in designing the computational strategies for multimodal representation, classification and retrieval.
 They share  a fundamental problem with  many other brain related fields such as clinical diagnosis of  mental diseases.  That is how to reliably locate the relevant brain regions corresponding to either external multimedia stimulus or certain mental disease. Via modern brain imaging techniques such as functional magnetic resonance imaging (fMRI), electroencephalography (EEG) and magnetoencephalography (MEG), ones can  image the structure, function or pharmacology of the nervous system in a non-invasive way. Recently, learning from neuroimaging data, as a kind of pattern recognition,  has led to impressive results \cite{Haxby01MVA}, such as  guessing which image or video a subject is looking at
 or where are the most activated brain regions when one responses to multimedia stimulus in brain disorders. {In many such applications, 
the commonly and reproducibly selected brain areas can be considered as the potential biomarkers \cite{Hilario08biomarker,Li09SparseVoxelSelection,Guyon03Introduction,Blum1997245}.  Without reliable, reproducible results, no study in generally can effectively contribute to scientific knowledge \cite{Bennett10reliableFMRI}. }

Discovering the  discriminative brain regions corresponding to  certain mental decease or activated brain regions corresponding to certain
external stimuli, also called support identification,  is a typical  feature selection problem in the context of pattern recognition. 
%
%
%
%
However, due to the high dimensionality of feature space and relatively few samples, it is quite a challenging task to accurately and stably estimate discriminative voxels  \cite{buhlmann2011statistics}. Many existing  multi-variable pattern analysis methods often fail to provide a stable feature selection result, i.e., there exists the inherent significant
run to run variability in the decision space generated by the classifiers in terms of  even very small changes to the training set \cite{Anderson10critiqueMVPA}. This instability  is
   partially because classic multivariate feature selection methods aim at selecting a
minimum subset of features to construct a classifier of the best predictive accuracy and  often ignore
``stability" in the algorithm design \cite{Yu08denseFeatureGroups,He2010215,buhlmann2011controlling,Cover65Pattern,Anderson10critiqueMVPA,Raizada2010PFN}. 

Therefore, we will focus more on the stability and reliability of the feature selection results, for the sake  of scientific truth. {  Correspondingly, 
 our research will be based on multi-center neuroimage data analysis because multi-center data allows  to test the
reproducibility and consistency of the feature selection  results. In addition,  a multi-center  study offers several
advantaged over a single center study. For example, it has the potential to increase the number and diversity of subjects enrolled, and to include significant numbers of subjects from
rare clinical subgroups \cite{Mueller10how}. While  inter-center variability brings great challenges for consistent feature selection results, it can also be used to test the performance of different algorithms. 
Finally, we would like to emphasize that in line with the trend towards Big Data in brain study, major advances have been made in the availability
of shared neuroimaging data. There have existed more than
$8,000$ shared MRI (magnetic resonance imaging) data sets available online \cite{Poldrack14neuroimaging}. The quantitative analysis of these multicenter data sets
 from related magnetic resonance imaging experiments has the potential to significantly accelerate progress in brain
mapping. }

In order to further increase the trustworthiness of the discovered biomarkers, we consider both functional MRI (fMRI, for short) and structural MRI (sMRI, for short). While the results from functional imaging are increasingly being
submitted as evidence into the legal systems of many nations,  
the reliability of the evidence from fMRI still needs to further proved \cite{Bennett10reliableFMRI}. Therefore, we also consider the structural MRI data, which is generally considered to be less dynamically changed. The biomarker results from both fMRI and  sMRI will be considered together to give a better and more convincing understanding of the decease, i.e. Autism Spectrum Disorder (ASD) in this paper \cite{Serres14multimodals,Wainer14}.


{  To our best knowledge, most of existing algorithms fail to overcome  this difficulty caused by data variation. For example, 
%
the univariate approaches have been popular for finding the discriminative regions
due to their being directly
testable, easily interpretable, and computationally tractable. However, it is not quite robust to the data perturbation possibly caused by the between-center variation.  In addition,  there exists a strong demand  in various multivariate approaches in
recent years \cite{Haxby14Neural}, because the brain representation
is intrinsically multivariate. However, as mentioned before, many multivariate methods often ignore
``stability" in terms of feature selection. }

  {We aim to achieve a consistent feature selection result across different sites.  Correspondingly,  
%
%
%
%
we will revisit  our recently proposed concept of ``randomized structural sparsity" and  apply it to the multi-center data analysis. We will verify the stability of the resulted algorithm in terms of the generalization and reproducibility  of the selected features via the multi-center data \cite{yu2013}. 
In particularly, we will
%
   check whether a consistent feature selection result can be obtained across different centers, and  this consistency will support the reliability of the selected features as the potential true biomarker of the associated disease, i.e. ASD here.} 

\subsection{Advantages and Limitations of Sparsity Applied to Nueroimaging}
{While most of existing efforts to deal with multi-center data is to  merge the data from different sites into a pool  and analyze it as a whole, like a big single center, we  consider each of multiple centers individually \cite{Martino13MP}. } In particular, we consider each of the centers separately and apply the feature selection method to obtain the potential biomarker. Then we  compare the results from different centers to see whether they are consistent to a certain degree. While this seems to be a natural and better way to validate  the reproducibility of the final selected biomarkers than only considering a  single center data, it brings a great challenge to the feature selection method, which is required to be stable due to the possible high data variability across different centers. Now we first review some commonly used feature selection methods in the following parts.

We consider to  identify the discriminative brain voxels from the given labeled training MRI data in context of the supervise learning. 
While the  classification problem is mainly considered, the regression problem can be treated in a similar way.  The linear model is widely  used for feature selection as follows.
  \begin{equation} \label{eq:linearmodel}
 \mathbf{y} = \mathbf{X}\mathbf{w} + \epsilon
 \end{equation}
 where $\mathbf{y} \in \mathbb{R}^{n\times 1}$ is the binary classification  information and $\mathbf{X} \in \mathbb{R}^{n\times p}$ is the given training voxel-wise MRI data and $\mathbf{w} \in \mathbb{R}^{p\times 1}$ is the unknown weights reflecting the
 degree of importance of each voxel.  Identification of discriminative voxels  is based on the values of the weight vector $\mathbf{w}$ and their importance is  usually proportional to the absolute values of the components. 


Considering that the common challenge in this field is the curse of dimensionality $p \gg n$. 
 sparsity makes much sense that  the most discriminative activated voxels are only
a small portion of the total brain voxels \cite{Yamashita08SparseEstimation}. However, 
sparsity alone is not
sufficient for making reasonable and stable inferences, in cases of very high voxel space and small number of training samples. In such cases, the plain sparse learning models such as the $\ell_1$-norm regularized model also called LASSO \cite{tibshirani1996regression},  often provide
overly sparse and hard to interpret solutions 
\cite{Rasmussen2012SparsityFMRI}. 
Thus we need to extend the plain sparse learning model to incorporate some important structural features of brain imaging data in order to achieve a more stable, reliable and interpretable support identification results.

\subsection{Existing  Extensions of the Plain Sparse Model}

Functional segregation of local brain areas that differ in their anatomy and physiology contrasts sharply with their global integration during perception and behavior \cite{Tononi94segregation,Xiang:2009}, and   
sets of discriminative or activated voxels between different brain states are expected to form distributed and localized  areas \cite{Chu02featureselection,Tononi94segregation,LimbrickOldfield20121230}.  
Therefore,  two common hypothesis have been made for MRI data analysis. One is the sparsity: the discriminative or activated voxels implied in the classification task or an external stimulus are only a small portion of the total; the other is
 compact structure: these voxels are often grouped into several distributed  clusters because those within a cluster have similar behaviors. 
 Notice that the plain sparse learning model based on  the $\ell_1$ regularization is only making use of the first sparsity hypothesis and discard the structure related assumption. 

Elastic Net \cite{Zou05ElsticNet} tries to make use of the voxel correlation by adding an $\ell_2$ regularization or called Tikhonov regularization to the classical $\ell_1$ penalty. The strict convexity due to the help of the added $\ell_2$ term helps to select all together a group of redundant features or none of them in what is known as the grouping effect \cite{RyaliCSM12L1L2Network,Ryali10ElsticNetLogistic}. 
 {Recently, Total-Variation (TV) penalization was added to both $\ell_1$ penalization for voxel selection  \cite{Gramfort:2013:IPR:2552484.2552506}. The TV penalization is used to make use of the assumption that the activations are spatially correlated and the weights of the voxels are close to piece-wise constant. } 


Furthermore, structured sparsity models are proposed  to  more explicitly  make use of the segregation and integration of the brain  \cite{Bach12StructuredSparsity}, in order to extend the well-known plain $\ell_1$ models. They enforce more structured
constraints on the solution, 
such as the discriminative voxels are
grouped together in possibly few clusters \cite{Baldassarre12StructuredSparsityFMRI,Michael11TV,Ye12groupNonconvex,Liu10NIPSGroupeTree,Yuan13Overlapping,Jacob09GLO,Ng11classifier,Ye09L21}. 
In many cases, the parcellation information is not available beforehand, and therefore, ones can either use the anatomical regions as an approx \cite{Batmanghelich12Groupsparisty}, or  use the data driven methods to obtain the grouping information 
\cite{Michel2012SupervisedClustering,jenatton12multiscale}.
Notice that the kind of parcellation based on certain definition of homogeneousness  may not exactly match the ground truth of the discriminative voxels. However, most structural sparsity models tend to select either all voxels of a group or none of them, and fail to perform further refinements by selecting part of voxels of a group. 

%

%

Another important kind of feature selection methods for high dimensional data analysis is stability selection \cite{Meinshausen10StabilitySelection,Shah13StabilitySelection,Hastie09Ensemle}, which 
is  based on resamplings (bootstrapping would behave similarly) and  
aims to alleviate the disadvantage of the plain $\ell_1$ model, which either selected by chance non-informative regions, or neglected relevant
regions that provide duplicate or redundant classification information \cite{Mitchell04LearningDecodeCognitive,Li12SparseLocalization,Anderson10critiqueMVPA,Poldrack06Criti,Ye12SparseLearningSS,Cao2014Sparse,Ryali20123852}.
One main advantages of stability selection is the finite sample control of false positives. It also makes the  choice of sparsity penalty parameter  much less critical or sensitive \cite{buhlmann2011statistics}. However, it fails to explicitly make use of
the assumption that these targeted voxels are often spatially contiguous and form into distributed brain regions.
{Correspondingly,  we proposed so called ``randomized structural sparsity" based feature selction algorithm in \cite{Wang14RSS}, by incorporating the idea of structural sparsity into stability selection.  
}

\subsection{Our focus and  contributions}
{In this paper, we consider to achieve a consistent and reproducible  feature selection result.  
 We revisit the ``randomized structural sparsity" based feature selection method we recently proposed in \cite{Wang14RSS}, where the finite sample control of false positives and false negatives are considered on a single-center study. Here we demonstrate its reproducibility and stability of the selected biomarkers verified through the multi-center data analysis. 
 We make a further analysis of this ``randomized structural sparsity" based algorithm, by considering the sensitivity of several key parameters of our algorithm such as the number of the clustering, the block size of block samplings, for the first time. In addition,  we proposed a simple but effective heuristic way for setting the threshold value, which is used to filter out the uninformative features. Finally, we consider both structural MRI and functional MRI modalities.  The revealed potential biomarkers from these two modals will be put together for  better understanding of the ASD than the single modality.}

The rest of the paper is organized as follows. In section \ref{Sec:Method}, we revisit our recent algorithm for stable voxel selection. In section \ref{Sec:Experiments}, we demonstrate the advantages of our algorithm  on the real multi-center neuroimage data in terms of better consistency of selected discriminative voxels  across different centers. 
In section \ref{Sec:conclusion}, a short summary of our work and some possible future research directions will be given. 

\section{Randomized Structural Sparsity Applied to Multi-Center Data Analysis} \label{Sec:Method}

\subsection{The Background and Motivation }

{We focus more on how to obtain a reliable support identification in context of many fields including mental decease or external stimuli including multimedia recognition. This is not a trivial task. In this paper, we revisit  the Randomized Structural Sparsity (RSS, for shot) based algorithm we proposed in \cite{Wang14RSS}, and aim to demonstrate that this algorithm can achieve a stable and consistent feature selection result even in the existence  of data variations.

MRI data is profoundly affected by experimental and methodological factors.
There is a high likelihood of introducing undesirable inter-site variability into the data, reducing the benefit of the multi-center design. Here we take this challenge of data variation among  multiple centers as a favor to demonstrate the stability and reliability of RSS. 
We empirically show that  most existing feature selection methods fail to obtain a consistent result because they are not able to deal with  the possibly relatively high data variability across different centers.}



\subsection{Review of Randomized Structural Sparsity}
We first review the Randomized Structural Sparsity based feature selection method. 
%
%
%
%
 Denote an MRI data matrix as $\mathbf{X} \in \mathbb{R}^{n\times p}$ where $n$ is the number of samples and $p$ is the number of voxels with $n \ll p$, and corresponding classification labels $\mathbf{y} \in \mathbb{R}^{n\times 1}$, where the binary classification and $\mathbf{y}_i \in \{1,-1\}$ is considered here.  
We take the following sparse logistic regression for classification as the example  to describe the main idea of our algorithm. 
\begin{eqnarray}\label{eq:L1}
\min_{\mathbf{w}}\|\mathbf{w}\|_1 + \lambda \sum_{i=1}^{n} \log(1+\exp(-\mathbf{y}_i(\mathbf{X}_i^{T} \mathbf{w}+c)))
\end{eqnarray}
where $\mathbf{X}_i$ denotes the $i$-th row of $\mathbf{X}\in\mathbb{R}^{n\times p}$; $\mathbf{y} \in \mathbb{R}^{n\times 1}$ is the labeling information containing the classification information of each row of $\mathbf{X}$; $\mathbf{w} \in \mathbb{R}^{p\times 1}$ is the weight vector for the voxels; $c$ is the intercept (scalar). The voxels corresponding to $\mathbf{w}_i$ of large absolute value will be considered as the discriminative voxels. 

Structured sparsity models beyond the plain $\ell_1$ models  have
been proposed to  incorporate  more structured
constraints on the solution \cite{Bach12StructuredSparsity,Li13ClusterGuided,Mairal13SFS}. 
The common way to make use of the clustering or grouping structure is the following group sparsity induced norm induced  model \cite{Bach12OSP}.
\begin{eqnarray}\label{eq:structuralL1}
\min_{\mathbf{w}}\sum_{g\in \mathcal{G}}\|\mathbf{w}_g\|_2 +  \lambda \sum_{i=1}^{n} \log(1+\exp(-\mathbf{y}_i(\mathbf{w}^{T}\mathbf{X}_i+c))),
\end{eqnarray}
where $\mathcal{G}$ is the grouping information. 
However, the obtaining of appropriate $\mathcal{G}$ might be difficult in practice, and the final results might be too biased by the grouping information $\mathcal{G}$.   
Recently, stability selection \cite{Meinshausen10StabilitySelection} based on the plain $\ell_1$ model has been widely applied \cite{Ye12SparseLearningSS,Cao2014Sparse,Ryali20123852}, due to its finite sample
control of the false positives.  In addition, it makes the choice of the regularization parameter insensitive.
However, it fails to make use of the prior structural information of the discriminative voxels,  and may result into a big false negative rate in order to keep low false positive rate.

Therefore, we proposed ``randomized structural sparsity" in \cite{Wang14RSS}, which  incorporates the spatial structural knowledge of voxels into the stability selection framework. It aims to  achieve  low false positive rate and false negative rate simultaneously.  In the paper, we will further reveal its stability and reliability via the multi-center data analysis, i.e. a consistent biomarker can be found across different sites due to its robustness to relatively small data perturbation. It is of important significance for the reliable scientific truth.
While ``randomized structural sparsity" is a general concept, it has a specific implementation  for voxel-wise MRI data anlaysis,  named ``Constrained Block Subsampling". 

 ``Constrained Block Subsampling"  prefers to using the block subsampling based stability selection \cite{Beinrucker12ExtensionSS}, rather than the original reweighting based stability selection \cite{Meinshausen10StabilitySelection}. 
%
Specifically, for the training data matrix $\mathbf{X}\in\mathbb{R}^{n\times p}$, subsampling based stability selection consists in applying
the baseline, i.e. the pure $\ell_1$ regularization model such as (\ref{eq:L1}), to random submatrices of $\mathbf{X}$ of size $[n/L] \times [p/V]$, where $[]$ is  the round off to the nearest integer number, and returning those features having the largest selection frequency. The original stability selection \cite{Meinshausen10StabilitySelection} can be roughly considered as a special case of it, where $L=2$ and $V=1$, except that the original stability selection \cite{Meinshausen10StabilitySelection} reweighs each feature (voxel, here) by a random weight uniformaly sampled in $[\alpha,1]$ where $\alpha$ is a positive number, and subsampling can be intuitively seen as a more crude version
by simply dropping out randomly a large part of the
features \cite{Beinrucker12ExtensionSS}. 
%
 %
 We further proposed to make use of the block subsampling \cite{Lahiri99Blockbootstrap},
which aims to replicate the correlation by subsampling  blocks of voxels instead of scattered voxels. 
The intuition  of ``blocking" exists in the assumption that the voxels are partitioned into spatially contiguous homogeneous subgroups, though possibly distributed. 
Moreover, we  incorporate the  parcelling information of  the brain into block subsampling, resulting in the so called ``constrained block subsampling'', where the ``constrained" means that the parcelling information will be respected.  The kind of partition information can be based on either the prior anatomical knowledge of brain partition \cite{TzourioMazoyer2002AnatomicalParcellation}, or the clustering results based on the MRI data. In particular, for each cluster $g \in \mathcal{G}$, it may consist of either only one or several distributed localized brain regions or called partitions, because  a cluster based on ``homogeneousness" could be disrupted into several different brain areas \cite{TzourioMazoyer2002AnatomicalParcellation}. 
Correspondingly,   
the selected voxels from the same cluster, after the common block subsamplings, will be considered as a subgroup, i.e.,  
 the chosen voxels lying in a  cluster $g \in \mathcal{G}$ are noted as  a set $g' \subseteq g$. Furthermore, in order to make sure that those small clusters can also be sampled during the block subsampling, we borrow some idea of ``proportionate stratified sampling" \cite{Sarndal03stratified,Devries86}, i.e. the same sampling fraction is used within each partition, in order  to reduce the false negatives, especially when the sizes of different partitions are of quite a range. 
%
%
This way, we obtain the following group-sparsity based recovery model. 
\begin{eqnarray}\label{eq:structuralL1Subsampling}
\min_{\mathbf{w}'}\sum_{g' \subset g\in \mathcal{G}}\|\mathbf{w'}_{g'}\|_2 +  \lambda \sum_{i \in \mathcal{J}} \log(1+\exp(-\mathbf{y}_i(\mathbf{w'}^{T}\mathbf{X}'_i+c)))
\end{eqnarray}
where $\mathbf{w}'$ and $\mathbf{X'}$ are corresponding parts of $\mathbf{w}$ and $\mathbf{X}$, respectively,  based on the selected voxels during the subsampling, and  $\mathcal{G}$ is a predefined or estimated partitions of the brain. $\mathcal{J}$ is the set of the indices of the selected samples of the current subsamling.

While  ``constrained block subsamplings"  respects the prior knowledge $\mathcal{G}$, it also provides the flexibility that discovered discriminative regions can be of any shape. The final selected voxels from a cluster can be only part of  it, because   the randomness of the block subsampling in the different iterations of the stability selection procedure, makes the selection frequency score
 be able to outline structures of the true discriminative regions.  This kind of flexibility is of importance because  the group information  might not exactly reflect the true shapes of the discriminative brain regions. The grouping information  is usually obtained via certain clustering algorithms which often only depend on the correlation information of voxels.

For the specific multi-center data analysis, the inherent variability between different centers \cite{Friedman2006varability,Friedman06JMRI} requires our algorithm to pay more attention to the reduction of variance of feature selection results, though it may result in certain  bias increasing, due to the bias-variance dilemma or bias-variance tradeoff  \cite{Geman:1992:NNB}.
%
In general, we would like to pay a little bias to save a lot of variance. Considering dimensionality reduction can decrease variance by simplifying models \cite{James13StasiticalLearning},   
we still use the   ``averaging" idea \cite{Varoquaux12Clustering} applied to  (\ref{eq:structuralL1Subsampling}), 
because \cite{Park01042007} has proved that when the variables or features were positively correlated, their average was a strong feature, yielding a fit with lower variance than the individual variables. Specifically,
by averaging the voxels picked by the block subsampling lying in the same cluster as a single super-voxel,  the model  (\ref{eq:structuralL1Subsampling}),  can be further reduced to the following reduced dimensional version
\begin{eqnarray}\label{eq:everaged}
\min_{\tilde{\mathbf{w}}}\sum_{g' \subset g\in \mathcal{G}}|\tilde{\mathbf{w}}_{g'}| +  \lambda \sum_{i \in \mathcal{J}} \log(1+\exp(-y_i(\tilde{\mathbf{w}}^{T}\mathbf{\tilde{X}}_i+c)))
\end{eqnarray}
where $\tilde{\mathbf{w}} \in \mathbb{R}^q$, and $q$ is the number of clusters. $\tilde{\mathbf{w}}_{g'} $ is an average of voxels in the subset $g'$ of cluster $g\in \mathcal{G}$, and $\mathbf{\tilde{X}}\in \mathbb{R}^{[\alpha n] \times p}$ is the corresponding averaged $\mathbf{X}$.
If the $j$-th column of $\tilde{\mathbf{X}}$ is selected due to the large magnitude of $\tilde{\mathbf{w}}_j$, then its represented picked blocked voxels lying in the group $g^{(j)} \in \mathcal{G}$ ($j=1,2,\ldots, q$) of $\mathbf{X}$ are all counted to be selected, in 
the non-clustered
space, and its corresponding score $\mathbf{s}_i$ will be updated ($i=1,2,\ldots,p$). Notice that the ``averaging" of sumsampling is more than a simple spatial smoothing, due to different sumsampling results of different stability selection iterations. Therefore, the boundaries of the detected discriminative regions can be still trusted to certain accuracy. 

\section{Numerical Experiments} \label{Sec:Experiments}
In this paper, we will demonstrate
the stability and reliability of the RSS based support identification method. This point will be empirically verified  based on the multi-center data, which is of data variation among different centers. In this paper, the data comes from
  the Autism Brain Imaging Data Exchange (ABIDE)- a grassroots consortium aggregating and openly sharing $1112$
existing resting-state functional magnetic resonance imaging (R-fMRI) data sets with corresponding structural MRI (sMRI) and phenotypic
information from 539 individuals with ASDs and 573 age-matched typical controls (TCs; 7-64 years) (http://fcon\_1000.projects.
nitrc.org/indi/abide/).  Then we follow the sample selection principle in \cite{Martino13MP} 
get the final data for $763$
individuals (ASDs=360; TCs=403) from $14$ centers. 

{We compare with two sample t-test, $\ell_2$ logistic regression, $\ell_2$ SVM, TV-L1 \cite{Gramfort:2013:IPR:2552484.2552506}, $\ell_1$ logistic regression and randomized $\ell_1$ logistic regression (stability selection based on $\ell_1$ logistic regression). Two sample t-test and $\ell_2$-SVM are implemented as internal functions of MATLAB. The python code of TV-L1 is provided by Prof. Alexandre Gramfort of Telecom ParisTech and it is  under integration in the Nilearn package. Most of the other  algorithms have been implemented in LIBLINEAR \cite{REF08a} or SLEP (Sparse Learning with Efficient Projections) software \cite{Liu09:SLEP:manual}.} 

\subsection{Image preprocessing}
R-fMRI scans were preprocessed with 【SPM】(\url{www.fil.ion.ucl.ac.uk/spm/}). Image preprocessing steps
included slice-timing and motion correction, smoothing, correted R-fMRI
measures were normalized to Montreal Neurological Institute (MNI) 152
stereotactic space (3$mm^3$ isotropic) with linear and non-linear registrations.Linear detrend and temporal filter (0.01-0.08Hz) were then applied on the normalized images. For each participant, we generate the following voxel-wise regional metrics: Amplitude of Low Frequency Fluctuations (ALFF) \cite{YuFeng2007ALFF}.  ALFF is defined as the mean square root of the power spectrum
density over the low frequency band (usually 0.01,0.08 Hz). After
an individual ALFF map is obtained, 
zALFF is obtained by performing  a standard z-transformation  on the voxels of individual ALFF map within a specific mask (gray matter mask). 

For structural MRI (3DT1, here), we perform Voxel-based Morphometry (VBM), which involves a voxel-wise comparison of the local concentration of gray matter between two groups of subjects \cite{Ashburner20008VBM}.  All the skull-stripped and reoriented images were spatially normalized to the Montreal Neurological Institute (MNI) space by minimizing the residual sum of squared differences between structural MRI and the ICBM 152 template image. The data were then resampled to $3\times 3\times 3mm^3$. All these images were segmented into GM (grey matter), WM (white matter) and CSF (cerebrospinal fluid) using the unified segmentation algorithm with incorporated bias correction. GM images were then smoothed with an 8-mm smoothing kernel.  In this paper, we use the preprocessed VBM data as a main feature to test our algorithm.



\subsection{Settings of Algorithms}
 For our algorithm, 
 we use the  SLEP \cite{Liu09:SLEP:manual} software to efficiently solve the model (\ref{eq:everaged}), which in fact  a common $\ell_1$ model.
For the selection of regularization parameters of the involved multivariate methods except our method,  cross validation is used.
{ We set the subsampling rates are $0.5$ and $0.1$ for samples and voxels, respectivley. The resampling times is  $200$. These settings are following the common stability selection default settings \cite{Meinshausen10StabilitySelection}. }


 {
  The internal k-means function of MATLAB is used for clustering. The default settings are used. For example, Squared Euclidean distance is adopted. One key parameter is the number of clusters, which is need to be prescribed.  The number of parameters are set according to the number of samples. For this multi-center data, each center has around $55$ samples. In general, we heuristically set the number of clusters between twice and $5$ times of the number of samples. We have tried $3$ different  numbers of clustering such as 116, 160 and 200, respectively,  in order to test how sensitive our final results are with this number. } 

{The block size is another key parameter of our algorithm.  It can be set according to the probably available prior knowledge of the size of the discriminative regions. In case of no such prior knowledge, we can give it a moderate value no larger than $10\times 10\times 10$. In the following subsection, we try different size of block from $3\times 3 \times 3$ to  $7 \times 7 \times 7$, which are considered to be in a reasonable range based on our experimental experiences. We can see that the final voxel selection result is not sensitive to the block size in this experiment. }

{In order to control the false positives, we need to set a threshold value to filter out uninformative voxels, i.e.,   voxels whose corresponding weights have smaller magnitude than this threshold will be considered as noisy voxels. 
The setting of a threshold value is quite difficult in general and may adopt different schemes in different situations. 
%
For two sample t-test method,  the number is generally determined by setting the p-value $ <0.05$ as significant level. 
For  the multivariate methods, 
%
while the cross-validation
method is widely used, it usually works well for the cases where there is a lot of training data available
and the prediction accuracy is the main concern. In addition, it often causes large false negative rate though a small false positive rate can be often achieved. 
} In this paper, we use a very simple but flexible way to set the threshold value.  It is a probability based method suggested in \cite{Li09SparseVoxelSelection}.  We repeat the description of the probability method in \cite{Li09SparseVoxelSelection} as follows.  
%
%
Specifically, considering that the entries of $\mathbf{w}$ are sparse,
we assume that the probability distribution of the entries of
$\mathbf{w}$ is Laplacian. Using all entries of $\mathbf{w}$ as samples, we estimate
the mean, the variance, and the inverse cumulative distribution
function F-1 of this Laplacian distribution. We then
define $R = \{i: |\mathbf{w}_i| > \theta, i = 1, ・ ・ ・,p\}$, where $\theta$ is chosen as
F-1 ($p_0$), $p_0$ is a given probability (e.g., $0.975$ or more restrictive $0.99$ in this paper). 
{  Unlike the single center data analysis, the threshold is not necessarily very  restrict because  the operation of intersection of the selected features between different sits helps remove false positives.  } Moreover,  we can gradually reduce the $p_0$ from a large value (for example, $0.99$) and check the corresponding selected voxels, which are the intersection of the selected voxels of each center. We have observed that when $p_0$ is reduced to a certain value (for example, 0.95), the selected voxels would  stay unchanged. In such cases, we have obtained a reasonable threshold. We will show this heuristic rule in the following description of the numerical results.

 \subsection{Evaluation Criteria}
We would like to  demonstrate that our method can achieve  a consistent feature selection result across different centers, due to  our  incorporation of  ideas of stability selection and structural sparsity.
%
%
 Since there is no ground truth for  evaluation, the confidence level of a selected relevant voxel depends on the number of centers where it is commonly selected. Since the feature selection algorithm is performed on each center  individually, a voxel appearing at the voxel selection results of many centers is more likely to be true informative voxel. Therefore, we will show and compare the consistent selected voxels among different centers of all these involved algorithms.

{ In addition, in order to further demonstrate that  these intersected regions discovered  by our algorithm are  stable, we  perform a false positive estimation scheme based on the permutation test and cross validation, which is proposed in \cite{Rondina14}. It  aims to calculate the ratio of false positives among all the finally selected voxels. 

Finally, we also present the prediction power of the selected voxels. The details are in the following subsection \ref{subsec:analysis}.} 

\subsection{Clustering Results and Feature Selection Results}
\begin{figure*}[!h]
    \centering
    \includegraphics[width=0.7\textwidth,height=0.18\textwidth]{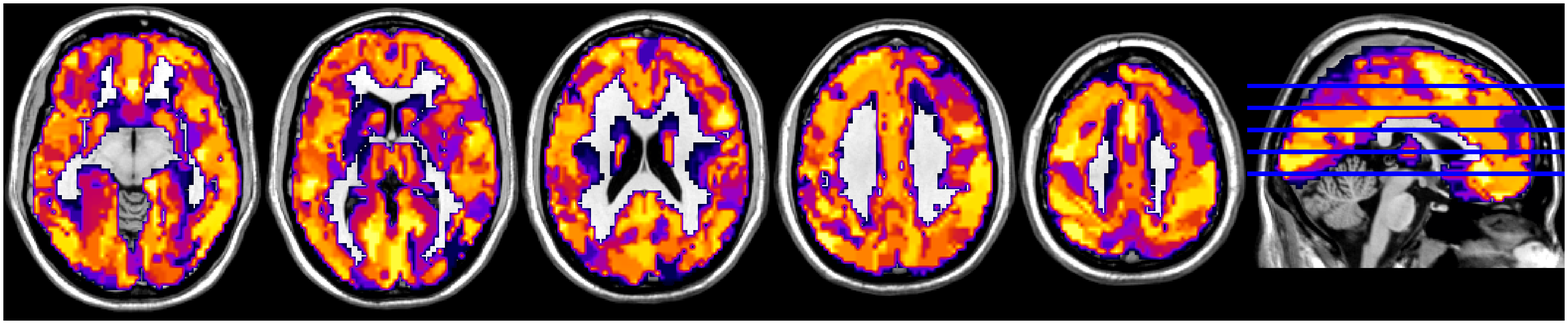}\vspace{0.1cm}
    \includegraphics[width=0.7\textwidth,height=0.18\textwidth]{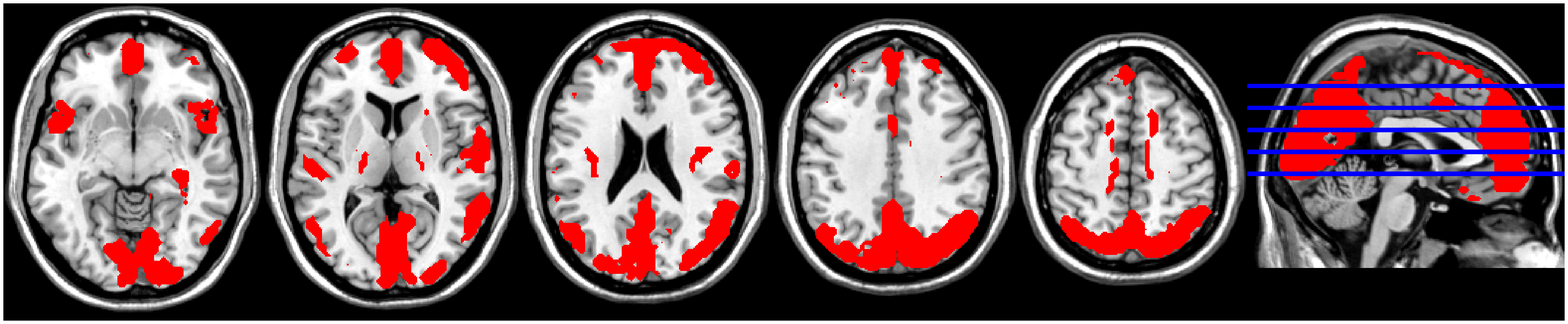}
    \vspace{-0.1cm}\caption{The upper row is the clustering result by K-means with the number of clusters being $200$, for the fMRI data of SDSU center. The lower row is the brain map of the selected voxels on the same center, with the threshold parameter $p_0=0.975$.
   }\label{Fig:Clustering_fMRI}
\end{figure*}

\begin{figure*}[!h]
    \centering
    \includegraphics[width=0.7\textwidth,height=0.18\textwidth]{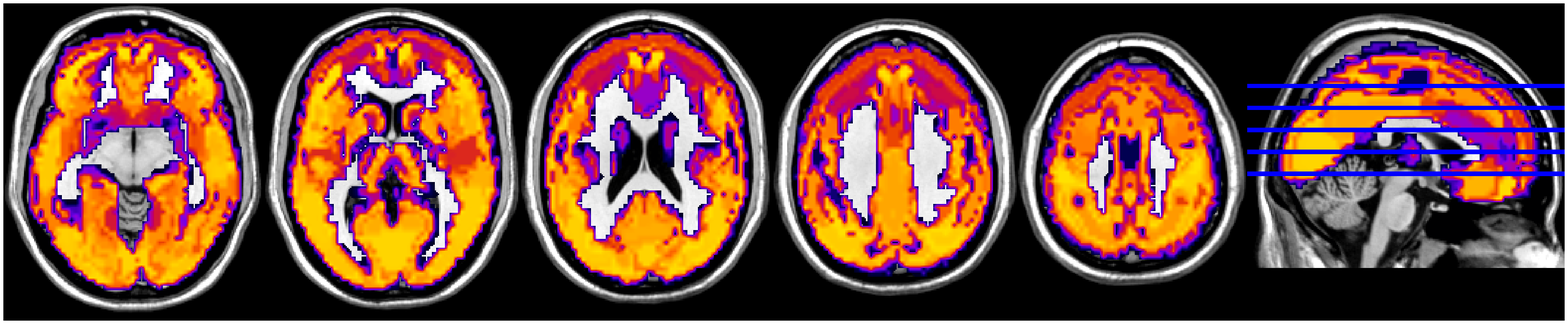}\vspace{0.1cm}
    \includegraphics[width=0.7\textwidth,height=0.18\textwidth]{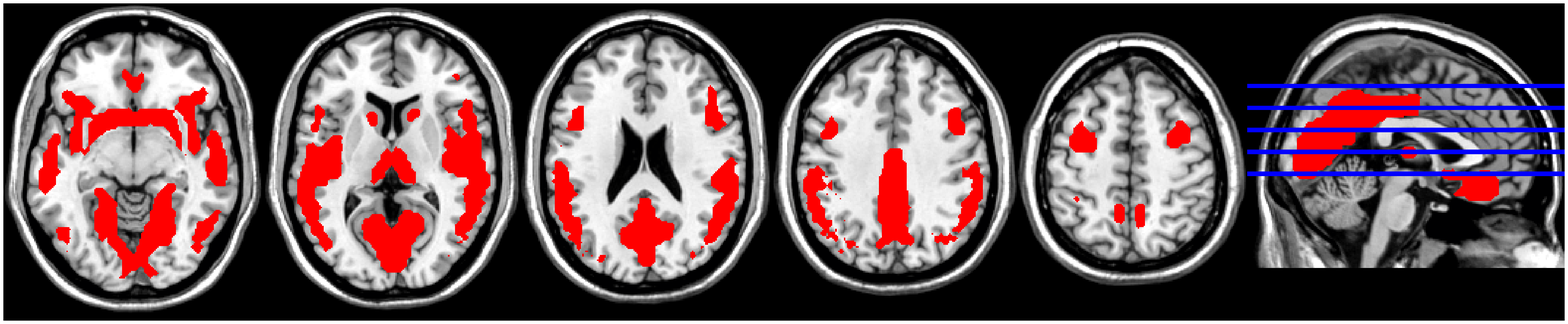}
    \vspace{-0.1cm}\caption{The upper row is the clustering Results by Kmeans with the number of clusters being $200$, for the sMRI data of NYU center. The lower row is the brain map of the  selected voxels on the same center, with the threshold parameter $p_0=0.975$.
   }\label{Fig:Clustering_sMRI}
\end{figure*}

{As mentioned before, the proposed RSS challenged the limitations of stability selection. The original stability selection  fails to explicitly make use of the assumption that these targeted voxels are often spatially contiguous and results in distributed brain regions. We used K-means to cluster the voxels into groups to model the spatial constraints. We need to emphasize that  the clusters resulted from K-means can not accurately model the underline homogeneity of the brain regions. However, due to the adoption of subsamplings, we can reduce the bias of the final feature selection results caused by K-means-based clustering.  Figure \ref{Fig:Clustering_fMRI} is the K-means clustering result and corresponding feature selection result on fMRI data of the SDSU center.
We can see the results of K-means clustering are reasonable and roughly match the prior  knowledge of brain segmentation and integration to some degree. In addition, while the clustering information helps reduce the variance, the final selected brain regions do not necessarily exactly follow the grouping based on the K-means clustering. This way, the possible bias caused by the clustering is expected to be reduced. We can also observe similar phenomenon from  sMRI data, as showed in  Figure \ref{Fig:Clustering_sMRI}.  }

\begin{figure*}[!h]
    \centering
    \includegraphics[width=0.8\textwidth,height=0.85\textwidth]{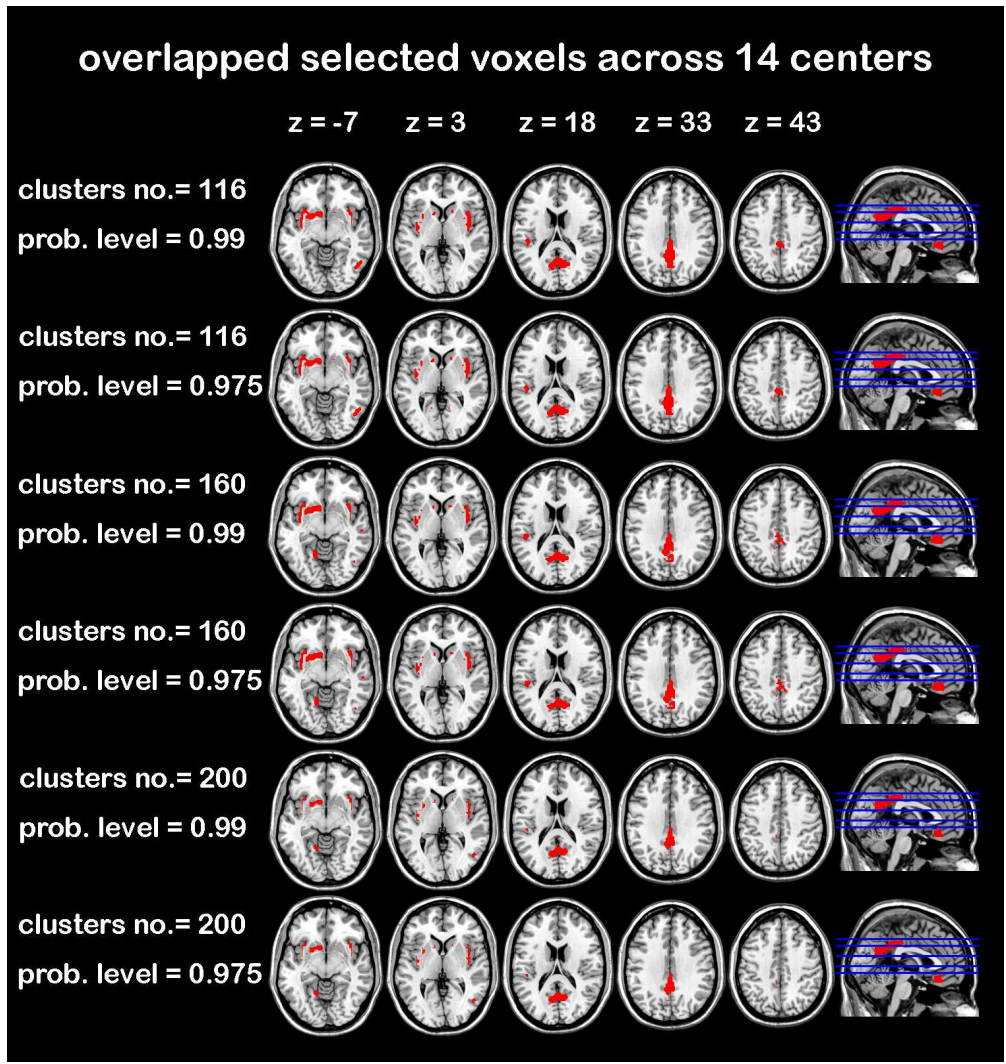}
    \vspace{-0.1cm}\caption{Selected overlapped voxels of our method shared by all 14 centers for the sMRI, i.e., 3DT1 data. We tested different settings of number of clusters (clusters no. ) and different $p_0$ (prob.level) to demonstrate the robustness of our algorithm.
   }\label{Fig:Our1-clusterignNumberSMRI}
\end{figure*}

\begin{figure*}[!h]
    \centering
    \includegraphics[width=0.8\textwidth,height=0.85\textwidth]{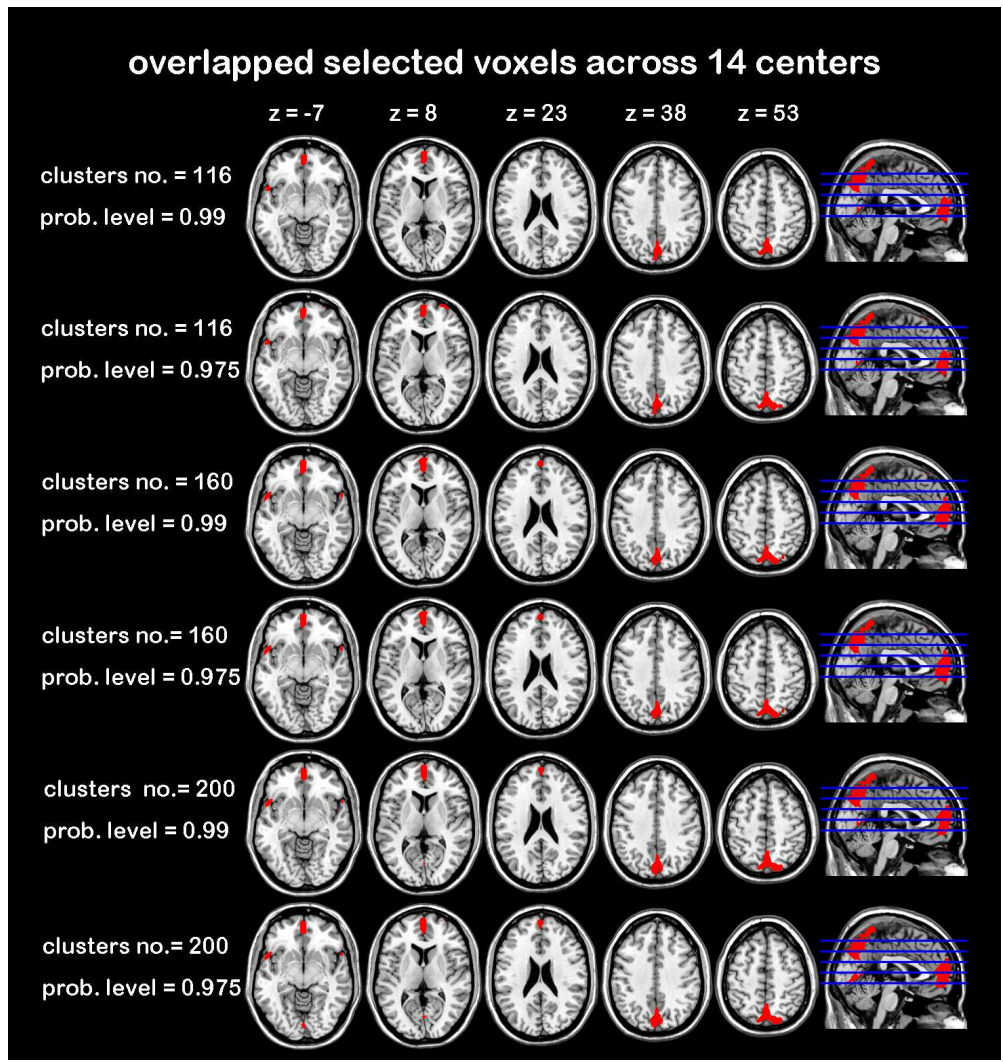}
    \vspace{-0.1cm}\caption{Selected overlapped voxels of our method shared by all 14 centers for the fMRI data. We tested different settings of number of clusters (clusters no. ) and different $p_0$ (prob.level) to demonstrate the robustness of our algorithm.
   }\label{Fig:Our1-clusterignNumberFMRI}
\end{figure*}

{  We have also tested $3$ different  numbers of clusters such as $116$, $160$ and $200$. Figure \ref{Fig:Our1-clusterignNumberSMRI} and Figure \ref{Fig:Our1-clusterignNumberFMRI} are the results of our algorithm applied to sMRI and R-fMRI data, respectively. We have observed that the final support identification results are not very  sensitive to the number of clusters. The revealed brain regions are the same, though they might be of slightly different size for different settings. }

\subsection{{ Feature Selection Results Corresponding to Different Block Sizes}}
\begin{figure*}[!h]
    \centering
    \includegraphics[width=0.8\textwidth,height=0.75\textwidth]{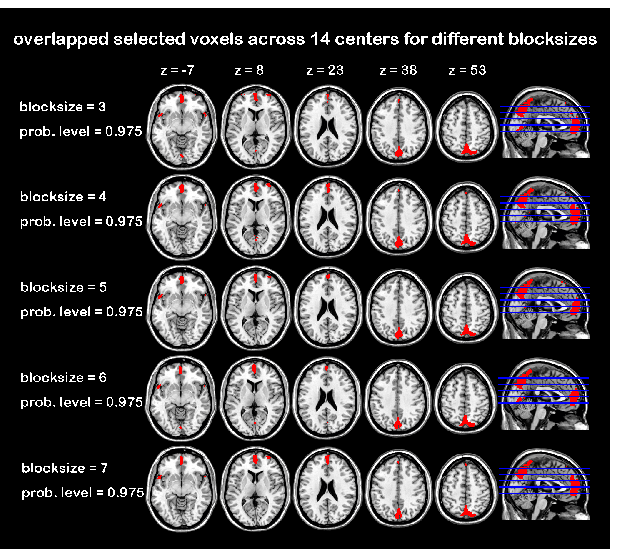}\vspace{0.1cm}
    \vspace{-0.1cm}\caption{Feature selection results of our algorithm corresponding to different block sizes for fMRI data.
   }\label{Fig:Our2-blocksize-fMRI}
\end{figure*}

\begin{figure*}[!h]
    \centering
    \includegraphics[width=0.8\textwidth,height=0.75\textwidth]{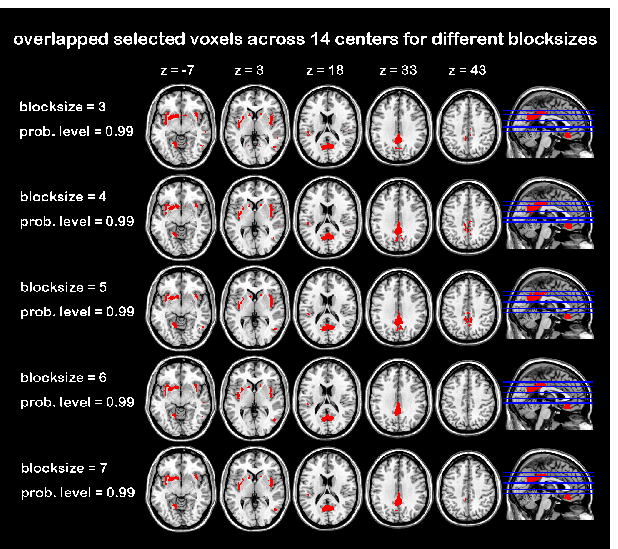}\vspace{0.1cm}
    \vspace{-0.1cm}\caption{Feature selection results of our algorithm corresponding to different block sizes for sMRI data
   }\label{Fig:Our-blocksize-sMRI}
\end{figure*}

{Now we are checking the robustness of our algorithm in terms of different block sizes when performing block subsampling. While we do not have the formula for  the optimal block size,  we show that the final feature selection result is not sensitive to the block size. We tested $5$ different block sizes, i.e. $3\times 3 \times 3=27$, $4 \times 4 \times 4=64$, $5  \times 5 \times =125$, $6 \times 6 \times 6=216$ and $7 \times 7 \times 7=343$. }  Figure \ref{Fig:Our2-blocksize-fMRI}  is the feature selection result when our algorithm is applied to the multi-center fMRI data set. We can see that  the results corresponding to different block sizes only have a small difference in term of the size of detected areas. Moreover, they all indicate the same  brain locations.  The similar conclusion can also be drawn for sMRI data, as suggested by Figure \ref{Fig:Our-blocksize-sMRI}.

\subsection{Comparison of Consistency of Results of Different Algorithms}

\begin{table}[!h]\tiny
\begin{tabular}{|l|l|l|l|l|l|l|l|l|l|l|l|l|l|l|l|l|l|l|l}
  \hline
Algorithms&ttest&	\multicolumn{2}{c|}{L2logistic}	&\multicolumn{2}{c|}{L2SVM}	&\multicolumn{2}{c|}{L1Logistic}		&\multicolumn{1}{c|}{TV-L1}&\multicolumn{2}{c|}{Randomized L1}&\multicolumn{6}{c|}{Our method}\\\hline
No. Clusters& &	\multicolumn{2}{c|}{}	&\multicolumn{2}{c|}{}	&\multicolumn{2}{c|}{}&\multicolumn{1}{c|}{}	&\multicolumn{2}{c|}{}&\multicolumn{2}{c|}{116}&\multicolumn{2}{c|}{160}&\multicolumn{2}{c|}{200}\\\hline
prob.level &	& 0.99    &0.975	& 0.99&0.975	& 0.99&0.975& & 0.99&0.975&0.99&0.975& 0.99&0.975& 0.99&0.975\\\hline												KKI	&	    1303&		10873&	16438&		10454&	16337&	12	&17&  8290&	8491&	14274&			8583&	8583&	8459&	8459	&8146	&8146\\\hline
Leuven&		1590&		10440&	15426&		10788&	16517&	51	&51	&	6963&8177&	13266&			7699	&7699&	7753	&7753	&8487&	8487\\\hline
MaxMun	&	1983&		10456&	15511&		11537&	17568&	63&	64	&	7492& 8250&	13145&			8105&	8105	&7975&	7975	&8395&	8395\\\hline
NYU	&	    2160&		10856&	16327&		10687&	16395&	422&455&	7405& 8539&	14572&			8109&	8109&	8108&	8108	&8334&	8334\\\hline
OHSU&		1643&		9508&	14368&		11158&	17148&	12&	12	&	6737&8009&	12853&			9007&	9007&	9105&	9105&	9693&	9693\\\hline
Olin&		958	&	    10259&	15196	&	11377&	17171&	1&	1	&	6332&8260&	13343&			9013&	9013&	8932&	8932&	8777&	8777\\\hline
Pitt&		954	&	    10764&	16084	&	11398&	17298&	8&	8	&	5341&8562&	14355&			8593&	8593&	8632&	8632&	8678&	8678\\\hline
SDSU&		1915&		10648&	15590	&	11970&	17985&	6&	6	&	7176&8628&	13362&			9766&	9766&	10477&	10477&	9753&	9753\\\hline
Standford&	2457&		10970&	16309	&	11061&	16869&	50&	50	&	10642&8655&	13981&			9063	&9063&	9033	&9033&	8636	&8636\\\hline
Trinity		&2423&		10246&	15355	&	11351&	17556&	9	&9	&	9370&8270&	12911&			8424	&8424	&8297&	8297&	8695&	8695\\\hline
UCLA	&	23643&		10659&	15921	&	10952&	16746&	178	&184&	7934&8324&	13729&			7547&	7547&	7807&	7807&	8092&	8092\\\hline
UM	&	    2587	&	10793&	16237	&	11571&	17607&	387	&413&	7043&8628&	14212&			8230&	8230&	8621&	8621&	8855&	8855\\\hline
USM	&	    8298&		10523&	15402	&	11577&	17181&	323	&330&	7714&8393&	13738&			7896&	7896&	7952&	7952&	7767&	7767\\\hline
Yale&		3801&		10462&	15617	&	11201&	17275&	2	&2	&	5749&8312&	12916&			9143&	9143&	8862&	8862&	8592&	8592\\\hline
overlapped(14)&		0&		0&	   2	&	0&	0&	0&	0&		0&0&	1&			1436&	1436	&1586&	1586&	1432	&1432\\\hline
overlapped(11)	&	0&		75&	531	&	0&	135&	0&	0&		0&177&	1664&			4557	&4557&	4674&	4674	&4838&	4838\\\hline
overlapped(9)	&	0&		703&	3430&		152&	2133&	0&	0&	0&	1101&	5924	&		6420&	6420&	6526&	6526&	6643	&6643\\\hline

\end{tabular}
\caption{Summary of Number of Selected Voxels for the multi-center sMRI data}\label{tab:summary2}
\end{table}

\begin{table*}[!h]\tiny
\begin{tabular}{|l|l|l|l|l|l|l|l|l|l|l|l|l|l|l|l|l|l|l|l}
  \hline
Algs&ttest&	\multicolumn{2}{c|}{L2logistic}	&\multicolumn{2}{c|}{L2SVM}	&\multicolumn{2}{c|}{L1Logistic}	& \multicolumn{1}{c|}{TVL1} &\multicolumn{2}{c|}{Randomized L1}&\multicolumn{9}{c|}{Our method}\\\hline
No. & &	\multicolumn{2}{c|}{}	&\multicolumn{2}{c|}{}	&\multicolumn{2}{c|}{}	&\multicolumn{1}{c|}{}&\multicolumn{2}{c|}{}&\multicolumn{3}{c|}{116}&\multicolumn{3}{c|}{160}&\multicolumn{3}{c|}{200}\\\hline
prob.level &	& 0.99    &0.975	& 0.99&0.975	& 0.99&0.975& & 0.99&0.975&0.99&0.975&0.96& 0.99&0.975& 0.96 & 0.99&0.975&0.96\\\hline												
KKI		&4347	&	10565&	15683&		11096&	16510&	249	&258  &5699  &	8380&	13286	&		9174&	9174& 9174&	7448	&7448&	7448 &7652	&7652&7652\\\hline		
Leuven	&2485	&	9884&	15090&		11709&	17578&	261&273	  &5868  &	8079&	13027	&		8293&	15210&	15210& 11964&	14248	&14248	&12720&14998&14998\\\hline		
MaxMun	&2819	&	9746&	14667&		11523&	17317&	277	&286 &5994&		8022&	12723	&		8872&	8872& 8872&	10148&	10148&	10148	&9964&9964&9964\\\hline		
NYU		&8339	&	9928&	15051&		11286&	16750&	653	&696	&5604&	8153&	13480	&		8290&	8290&8290&	8055&	8055&	8055	&8181&8181&8181\\\hline		
OHSU	&1785	&	9214&	13853&		11985&	17808&	239	&248	&5938&	7837&	12300	&		10831&	10831&	10831& 12604&	12604&	12604	&8077&12040&12040\\\hline		
Olin	&1532	&	10534&	15656&		11639&	17377&	189	&193	&5252&	8252&	13371	&		9618&	9618&	9618& 9515&	11491&	11491&	7981&11364&11364\\\hline		
Pitt	&1988	&	9966	&15057&		11847&	17811&	296	&321	&5863&	7999&	12901	&		8522&	8522& 8522&	7930&	8229&	8229&	8726&8726&8726\\\hline		
SDSU	&3539	&	10185&	15189	&	12117&	18201&	222	&229	&5938&	8635&	13629	&		11461&	11461& 11461&	10435&	10435&	10435&	9966&10412&10412\\\hline		
Standford	&	2486&10806&	16059	&	11772&	17467&	211	&217	&5853&	8529&	13836	&		9616&	9616& 9616&	9400&	9400&	9400&	11640&11640&11640\\\hline		
Trinity		&2742	&9874&	14815	&	12164&	18045&	349	&369	&5598&	8104&	12848	&		8385&	8385& 8385&	8964	&8964&	8964	&8447&8447&8447\\\hline		
UCLA		&1166	&9703&	14786	&	11818&	17501&	486	&500	&5516&	8046&	13191	&		8507&	10002&10002&	8895&	9896	&9896&	7920&9854&9854\\\hline		
UM		&3545	&10645& 	16123	&	11556&	17277&	508	&537	&5706&	8423&	13890	&		9417&	12500&12500&	10066&	12204&	12204&	8679&11344&11344\\\hline		
USM		&5238	&	9618&	14456	&	12003&	17765&	488	&511	&6006&	8183&	12954	&		7359&	7359&7359&	8892&	8892&	8892&	8786&8786&8786\\\hline		
Yale	&	2907&9112&  	13684	&	11314&	16652&	175	&180	&5874&	7609&	12026	&		11033&	11033&11033&	10598&	10598&	10598&	11630&11630&11630\\\hline		
OL(14)	&	0&	0&	2	&	0      & 	   1&	0	&0	&0	&1&	24	&	645 &	721&	721	& 688	&771&771	& 801&857&857\\\hline		
OL(11)	&	0&	104&	583&		4&	153  &	0	&0	&	0&158& 	1129& 2708&	3118&	3118&	2946&	3224&3224&	3131&3596&3596\\\hline		
OL(9)	&	0&	872&	3344&		153&2298	&0	&0	&	0&892&	3667& 4558&	4871&	4871&	4896&	5180&5180	&4968&5427&5427\\\hline	
\end{tabular}
\caption{Summary of Number of Selected Voxels for the multi-center fMRI data}\label{tab:summary1}
\end{table*}

  All the univariate and multivariate pattern feature selection methods can identified certain number of possible disease-related voxels for each center. The number has been listed in Tables  \ref{tab:summary2} and \ref{tab:summary1}, which corresponds to the results on the sMRI and fMRI, respectively.  Besides TV-L1, the multivariate methods mostly used  different $\theta$ values (denoted as prob.level), i.e. $0.99$, $0.975$ and even $0.96$ when applying the probability method to determine the selected voxels. For TV-L1, we have the third-party python software and we use its internal default way to set the threshold value, which is partially based on cross-validation.  The overlapped($S$), represents the number of the overlap of selected voxels across different centers, where $S$ means that at least $S$ centers share these selected voxels. As the $S$ increases, the number of overlapped voxels might decrease as expected.

\begin{figure*}
    \centering
    \includegraphics[width=0.8\textwidth,height=0.85\textwidth]{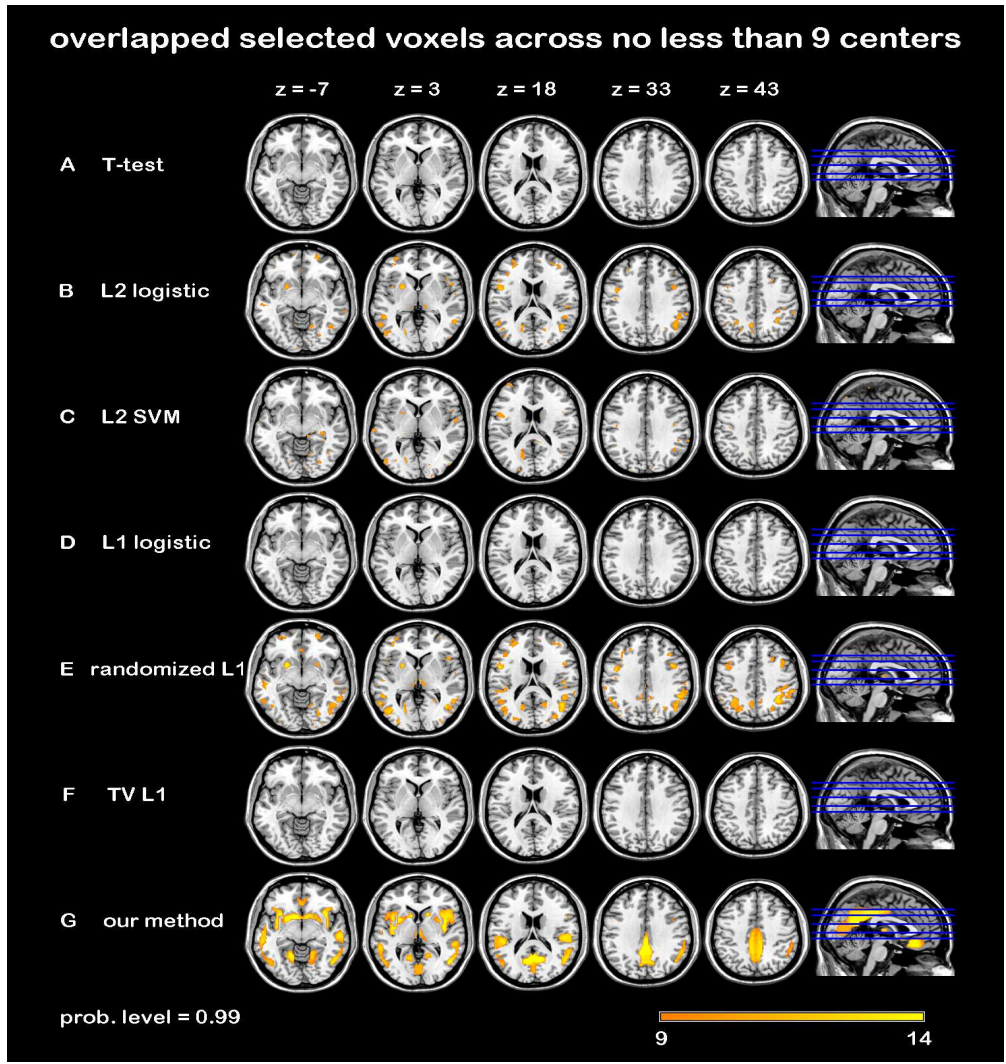}
    \vspace{-0.1cm}\caption{Selected overlapped voxels of our method shared by at least 9 centers for the sMRI data, where the probability value is 0.99. We compare the results by different methods. We can see our method is the best which can find out significantly interpretable discriminative structures. Randomized $\ell_1$ method and $\ell_2$ logistic regression are the second best and all the rest methods fail to find out consistent selected voxels among different centers.
   }\label{Fig:Compare9_T1}
\end{figure*}

\begin{figure*}
    \centering
    \includegraphics[width=0.8\textwidth,height=0.85\textwidth]{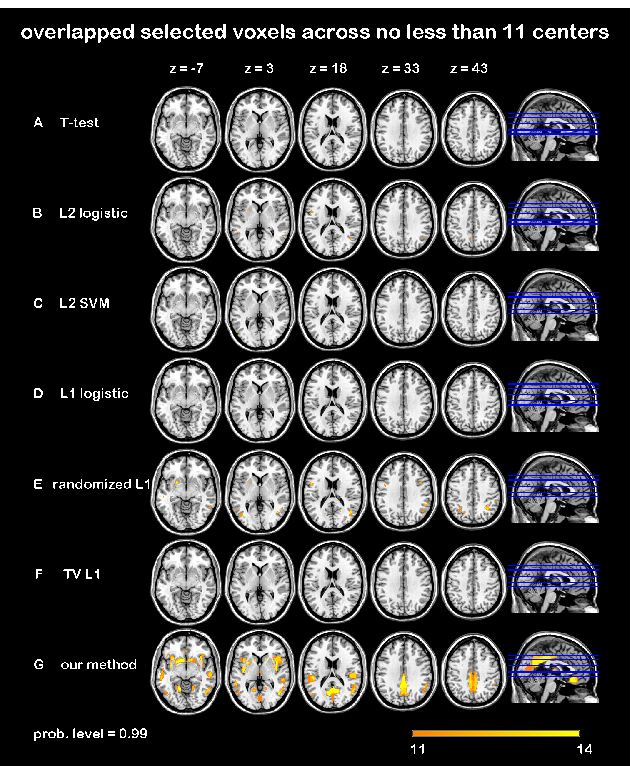}
    \vspace{-0.1cm}\caption{Selected overlapped voxels of our method shared by at least 11 centers for the sMRI data, where the probability value is 0.99. We compare the results by different methods. We can see our method is the best in terms of revealing   interpretable discriminative structures.
   }\label{Fig:Compare11_T1}
\end{figure*}

\begin{figure*}
    \centering
    \includegraphics[width=0.8\textwidth,height=0.85\textwidth]{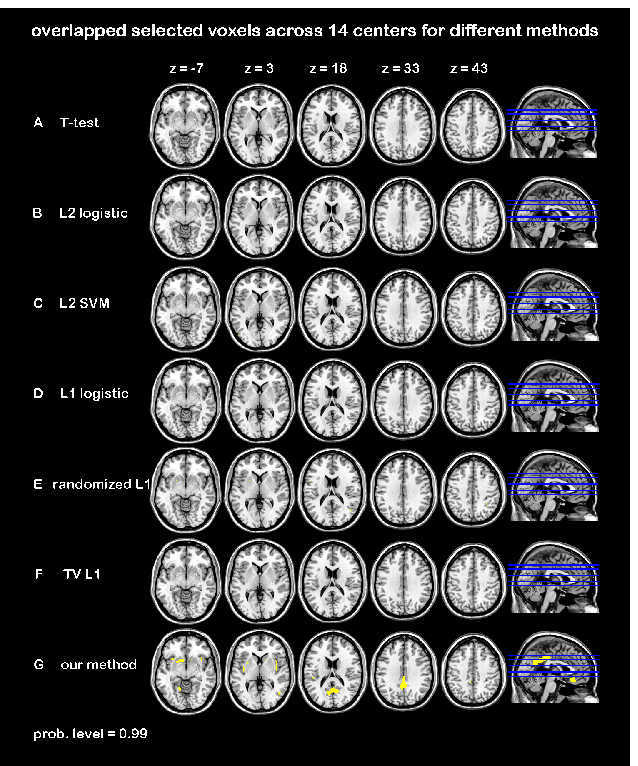}
    \vspace{-0.1cm}\caption{Selected overlapped voxels of our method shared by 14 centers for the sMRI data, where the probability value is 0.99. We compare the results by different methods. We can see our method is the  best in terms of revealing   interpretable discriminative structures.
   }\label{Fig:Compare14_T1}
\end{figure*}

\begin{figure*}
    \centering
    \includegraphics[width=0.8\textwidth,height=0.85\textwidth]{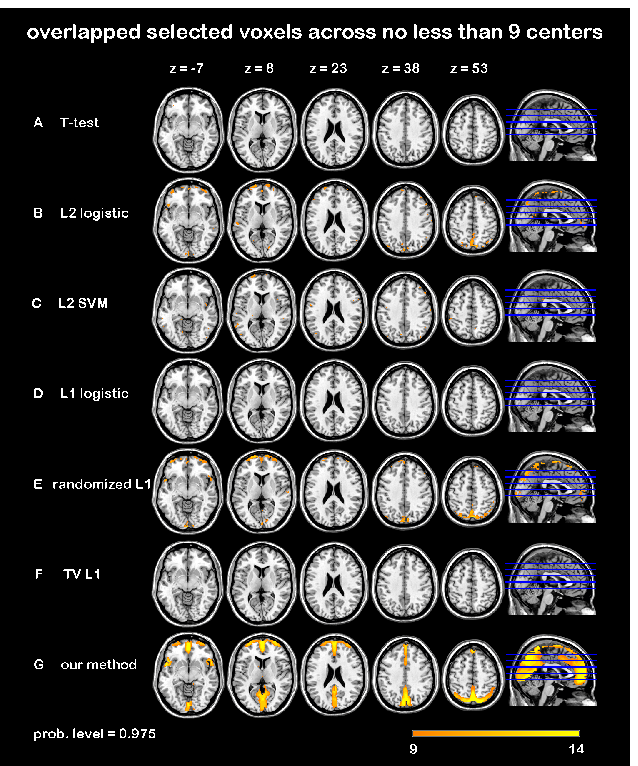}
    \vspace{-0.1cm}\caption{Selected overlapped voxels of our method shared by at least 9 centers for the fMRI data, where the probability value is 0.975. We compare the results by different methods. We can see our method is the best which can find out significantly interpretable discriminative structures. Randomized $\ell_1$ method and $\ell_2$ logistic regression are the second best and all the rest methods fail to find out consistent selected voxels among different centers.
   }\label{Fig:Compare9_fMRI}
\end{figure*}

\begin{figure*}
    \centering
    \includegraphics[width=0.8\textwidth,height=0.85\textwidth]{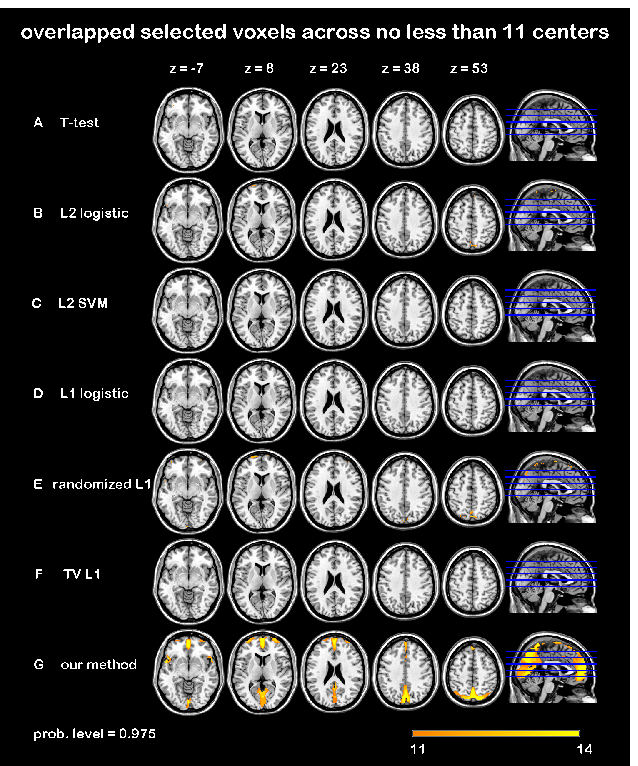}
    \vspace{-0.1cm}\caption{Selected overlapped voxels of our method shared by at least 11 centers for the fMRI data, where the probability value is 0.975. We compare the results by different methods. We can see our method is the best in terms of revealing   interpretable discriminative structures.
   }\label{Fig:Compare11_fMRI}
\end{figure*}

\begin{figure*}
    \centering
    \includegraphics[width=0.8\textwidth,height=0.85\textwidth]{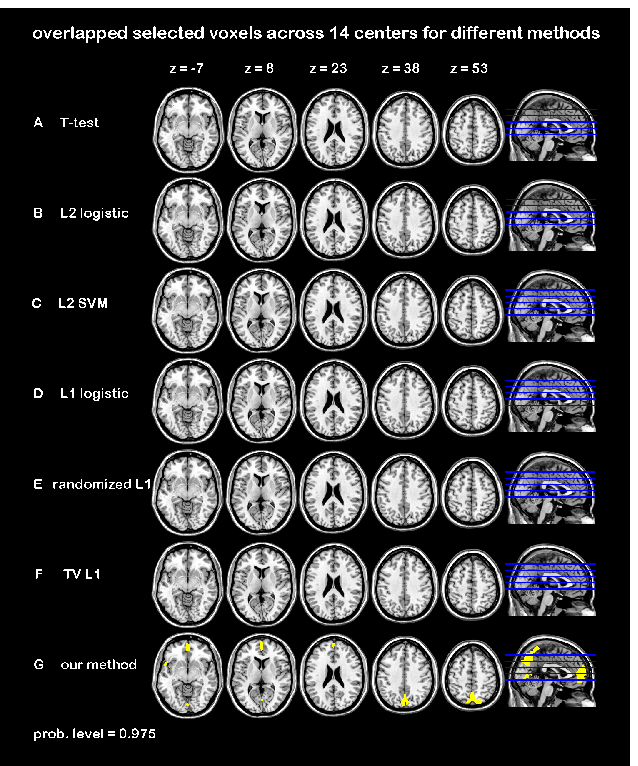}
    \vspace{-0.1cm}\caption{Selected overlapped voxels of our method shared by 14 centers for the fMRI data, where the probability value is 0.975. We compare the results by different methods. We can see our method is the best in terms of revealing   interpretable discriminative structures.
   }\label{Fig:Compare14_fMRI}
\end{figure*}

  The  variation of MRI data between different centers brings both challenges and benefits.
    On one hand, this kind of variation makes many existing state-of-the-art algorithms likely to select quite distinct voxels for different centers,   because they tend to select a
minimum subset of features to construct a classifier of the best predictive accuracy and  often ignore
``stability" of feature selection \cite{Yu08denseFeatureGroups,He2010215,buhlmann2011controlling}. Therefore, the results across different centers could be quite different and the overlapped parts are quite few. From  Tables  \ref{tab:summary2} and \ref{tab:summary1} , we can see that the overlapped voxels corresponding to the two sample t-test, L2-logistic, L2-SVM, TV-L1, and L1-Logistic are almost none when $S$ is larger than $11$. Notice that while  two sample t-test is widely believed to be a stable method for single center data analysis, it is not a good choice for multi-center data. TV-L1, as a recently popular feature selection method for the single center data analysis, does not work well for the multi-centere data analysis, partially due to the lack of stability scheme.   As expected, randomized $\ell_1$ logistic regression, as a method of stability selection, behavior  better than other alternatives. However, it is still much worse than our method, especially when $S$ is large, for example, $11$ or $14$, partially due to the failure of explicitly making use of the prior structural information of the discriminative voxels. For better comparison, We also plotted the selected voxels of different algorithms corresponding to $S=5, 9, 11, 14$, in Figures  \ref{Fig:Compare9_T1}, \ref{Fig:Compare11_T1} and \ref{Fig:Compare14_T1}  for the sMRI data, and  in Figures  \ref{Fig:Compare9_fMRI}, \ref{Fig:Compare11_fMRI} and \ref{Fig:Compare14_fMRI} for the fMRI data, respectively. 
We can see that our method can always achieve a consistent voxel selection result across different centers while the other state-of-the-arts mostly fail. 
%

  On the other hand, the variation of data between different centers can help us to verify the reliability of the selected discriminative voxels \cite{Friedman08retest}.   If these selected voxels are all shared by these involved centers, a better reliability of them is expected, even as the potential biomarker in the possible clinical diagnosis. Despite the computationally challenge, our algorithm incorporating the ideas of stability selection and structural sparsity successfully revealed significant number of the shared selected voxels.  

{Moreover, we would like to point out that  there exists a practical simple way to set the threshold value for feature selection for our algorithm. We have empirically observed that the sorted absolute values of the weights (from large to small) first gradually change then suddenly decreases dramatically and produce a gap. Usually we can set the threshold value at the location of the gap, and it is expected to achieve a good control of false positives. A similar  phenomenon has been observed and exploited in our previous related work about sparse signal recovery \cite{wang2010sparse}. Specifically, let us look at the Table \ref{tab:summary2}. For the two threshold values corresponding to probability levels $p_0=0.99$ and $p_0=0.975$, respectively,  the same discriminative voxels are obtained for our algorithm as each of the center. That is to say, the sorted weights of voxels indeed have a gap. Using this threshold value, we obtained an interpretable discriminative brain regions and we will give further analysis of these regions in the following subsection.  Table \ref{tab:summary1}  tells a similar story, except the gap is around the threshold values corresponding to the probability levels $p_0=0.975$ and $p_0=0.96$.  }




\subsection{{Subsequent Analysis of Feature Selection Results}} \label{subsec:analysis}
{We have observed that the overlap of selected voxels   by our algorithm is significantly higher. We believe that most of the selected voxols are true positives because the false positive usually are unlikely to repeat on every center. } Furthermore, we would like to further estimate the ratio of  false positives from another point of view based on permutation test and cross validation, as suggested in \cite{Rondina14}. The results are summarized in Table \ref{tab:summary-sMRI-fMRI}.  For sMRI, the estimated false positives ratios are all smaller than 3\%, for the overlap  of all the $14$ centers, for all the settings using different clustering numbers (116, 160, 200) and different probability levels (0.99 and 0.975). For fMRI, the estimated false positives ratios are slightly larger, but still smaller  than 6\%.  Notice that these numbers are only rough estimations for reference, not necessarily accurately reflecting the ground truth.  In addition, we provide the classification performances when using those identified voxels as features to construct a classifier. Notice  that for many alternative methods, they could not achieve consistent results if $S$ is large and so we have to reduce $S$ to $5$ or even $2$. For our method, we use $S=11$. When performing the classification, we pool all samples from $14$ centers together and we have in total $763$ samples. We randomly pick $700$ as the training samples and the rest $63$ as the test samples and we repeat this procedure for $100$ times. We use the $\ell_2$ logistic regression as the classifier. The classification accuracy is summarized in Tables \ref{tab:prediction-sMRI} and \ref{tab:prediction-fMRI} for sMRI data and fMRI data, respectively. While the selected voxels by our algorithm can achieve the best classification performance, they are generally not very high. A possible reason is that  a simple logistic classifier can not deal with the relatively high data variation of these samples from different centers and we might consider more sophisticated classifiers in the future.   


\begin{table}[!h]\tiny
\begin{tabular}{|l|l|l|l|l|l|l|l|l|l|l|l|l|l|l|l|l|l|l|l}
  \hline
&\multicolumn{6}{c|}{Our method}\\\hline
No. Clusters& \multicolumn{2}{c|}{116}&\multicolumn{2}{c|}{160}&\multicolumn{2}{c|}{200}\\\hline
prob.level &0.99&0.975& 0.99&0.975& 0.99&0.975\\\hline												
No. Totally Selected&					1436&	1436	&1586&	1586&	1432	&1432\\\hline
No. Estimated False Positives	&	25	& 25&	37&	37	&23&	23\\\hline
\end{tabular}\begin{tabular}{|l|l|l|l|l|l|l|l|l|l|l|l|l|l|l|l|l|l|l|l}
  \hline
&\multicolumn{6}{c|}{Our method}\\\hline
No. Clusters& \multicolumn{2}{c|}{116}&\multicolumn{2}{c|}{160}&\multicolumn{2}{c|}{200}\\\hline
prob.level &0.99&0.975& 0.99&0.975& 0.99&0.975\\\hline												
No. Totally Selected&					645&	721	& 688&	771&	801	&857\\\hline
No. Estimated False Positives	&	24	& 33&	27&	39	&45&	51\\\hline
\end{tabular}\caption{Number of Selected Voxels (across 14 centers) and Estimated False Positives. The left part corresponds to the  multi-center sMRI data and
 the right part corresponds to the  multi-center fMRI data}\label{tab:summary-sMRI-fMRI}
\end{table}

\begin{table}[!h]\tiny
\begin{tabular}{|l|l|l|l|l|l|l|l|l|l|l|l|l|l|l|l|l|l|l|l}
  \hline
ttest&l2logistic&L2SVm&L1Logistic&TVL1&Randomized L1&Ours\\\hline
overlapped(5)&overlapped(11)&overlapped(11)& overlapped(2)& overlapped(5) &overlapped(11)& overlapped(11)\\\hline
0.6046&0.6054&0.6121&0.6012&0.6113&0.5941&0.6332\\\hline
\end{tabular}\caption{Prediction power of the selected voxels for sMRI data}\label{tab:prediction-sMRI}
\end{table}

\begin{table}[!h]\tiny
\begin{tabular}{|l|l|l|l|l|l|l|l|l|l|l|l|l|l|l|l|l|l|l|l}
  \hline
ttest&l2logistic&L2SVm&L1Logistic&TVL1&Randomized L1&Ours\\\hline
overlapped(5)&overlapped(11)&overlapped(11)& overlapped(2)& overlapped(5) &overlapped(11)& overlapped(11)\\\hline
0.6062&0.5908&0.6145&0.5944&0.5997&0.5801&0.6283\\\hline
\end{tabular}\caption{Prediction power of the selected voxels for fMRI data}\label{tab:prediction-fMRI}
\end{table}

  %


\subsection{Discussion Based on Biomarkers from Multimodal Data}
We aim to show that these discovered biomarkers are interpretable and correspondingly support the reliability and stability of our algorithm. 
According to Figures \ref{Fig:Our1-clusterignNumberSMRI} and \ref{Fig:Our1-clusterignNumberFMRI} , we found the grey matter abnormalities in precuneus, posterior cingulate cortex, insular, middle temporal gyrus-R, fusiform gyrus, hippocampus, parahippocampal gyrus may have the potential be the related biomarkers of sMRI data(VBM) for distinguishing between the autism and the healthy controls. And with respect to the fMRI (zalff), medial prefrontal, precuneus, temporal pole, superior and medial frontal gyrus be the revealed regions as the discriminations of ASD.
The potential structural biomarker regions are mainly concentrated in the DMN (default mode network), such as precuneus and posterior cingulate cortex(PCC) which are regarded relating to primary responsible for autobiographical memory and self-reference processing. For example, precuneus and  PCC are primarily responsible for autobiographical memory and the self-reference processing. Significant abnormality in insula was also revealed, it is a critical component of SN(salience network) and would be involved in switching between central-executive and DMN \cite{Sridharan26082008}. And the SN is thought to regulate dynamic changes in other networks and the damage to the structure of SN should disrupt the regulation of associated networks such as DMN \cite{Bonnelle20032012}. Besides the brain regions mentioned above, some other regions such as fusiform gyrus, hippocampus and middle temporal gyrus-R， which are reported in previous studies \cite{Schumann14072004}. In our research, those overlapped regions we identified as discriminatied features, are found across all the 14 centers. Thus we can say that they may be the ASD-related stable biomarkers for the diagnoses and further analysis, according to the characters of our algorithm purposed and discussed before.
Similar to the structural data, the functional abnormalities of the ASD mainly lie in regions of medial prefrontal (mPFC), precuneus and temporal pole (above three are the components of the DMN) and anterior cingulate(ACC). mPFC is in charge of social cognition associated with self and others \cite{Gusnard27032001}.  In addition, the alterations of ACC, a part of the brain’s limbic system, may be relevant to the character of autism \cite{Bush2000215}. Accordingly, the consistent  overlapped features in ASD from multi-center data further confirmed the functional abnormalities in ASD \cite{Castelli02,Critchley00}.


\section{Conclusion and Future work} \label{Sec:conclusion}


%
In this paper, we consider a commonly existent problem in mental disease diagnosis or congtive science based on brain imaging, i.e. locating the discriminative or activitaed brain regions correspoing to the specific decease or certain external stimulus including looking at multimedia materials. We revisit our recently proposed algorithm which incorporates the ideas of stability selection and structural sparsity, and aims to show its abilitity to reliably and stably find out these intrested brain regions  despite the inherent data variations or noise.
We consider to  demonstrate our point via the multi-center MRI data analysis, which is also a promising research direction by itself.  Specifically, our algorithm helps us to discover a set of consistent selected voxels across different centers and these voxels are quite interpretable and can be possible biomarkers for the correspoinding medical diagnosis. Considering that most of existing algorithms fail to overcome the difficulty brought by the variation between centers and therefore are not able to get a consistent result, our algorithm also seems to be promising for the multi-center data analysis from the algorithmic point of view. However, some therotical analysis of the performance of our algorithm will be required, considering most of current study are more on the empirical side.
 We will further test the performance of our algorithm through  more kinds of multi-center data and feature types
 in the future work, epecially in the cognitive experiments including the study of brain responses to multimedia.
\section*{Acknowledgment}
Thanks to Prof. Alexandre Gramfort of Telecom ParisTech  for kindly providing us with  TV-L1 code, which is  under integration in the Nilearn package.

\bibliographystyle{IEEEtran}
\bibliography{GroupSparsity,Pattern_recognition,brainScience,StabilitySelection,multi-center}







\end{document}